\documentclass[runningheads]{llncs}
\usepackage{graphicx}
\begin{document}
\title{Detection of Abnormal Behavior with Self-Supervised Gaze Estimation}
%
%
\author{Suneung-Kim \and
Seong-Whan Lee}
\authorrunning{S. Kim et al.}
%
\institute{Department of Artificial Intelligence\\ Korea University, Seoul, Republic of Korea\\
\email{\{se\_kim, sw.lee\}@korea.ac.kr}}
\maketitle          
\begin{abstract}
Due to the recent outbreak of COVID-19, many classes, exams, and meetings have been conducted non-face-to-face. However, the foundation for video conferencing solutions is still insufficient. So this technology has become an important issue. In particular, these technologies are essential for non-face-to-face testing, and technology dissemination is urgent. In this paper, we present a single video conferencing solution using gaze estimation in preparation for these problems. Gaze is an important cue for the tasks such as analysis of human behavior. Hence, numerous studies have been proposed to solve gaze estimation using deep learning, which is one of the most prominent methods up to date. We use these gaze estimation methods to detect abnormal behavior of video conferencing participants. Our contribution is as follows. i) We find and apply the optimal network for the gaze estimation method and apply a self-supervised method to improve accuracy. ii) For anomaly detection, we present a new dataset that aggregates the values of a new gaze, head pose, etc. iii) We train newly created data on Multi Layer Perceptron (MLP) models to detect anomaly behavior based on deep learning. We demonstrate the robustness of our method through experiments.\\

\end{abstract}

\section{Introduction}

The recent outbreak of COVID-19 disease has led to many changes in the life system. Most of the face-to-face systems, such as education and meetings, are currently being conducted non-face-to-face. In addition, non-face-to-face testing is also increasing. For this reason, many places are interested in technology in video conferencing solutions, in particular cheating prevent systems for non-face-to-face testing are emerging as the important issue. Therefore, many studies are conducting to prevent cheating on non-face-to-face tests\cite{ref_1,ref_2,ref_3,ref_4}. However, the non-face-to-face cheating prevent system is still inappropriate to apply in real life because it is very difficult. Therefore, we present one framework that contributes to video conferencing solutions using gaze estimation algorithms to approach these problems.

\begin{figure}
\centering
\includegraphics[width=1\textwidth]{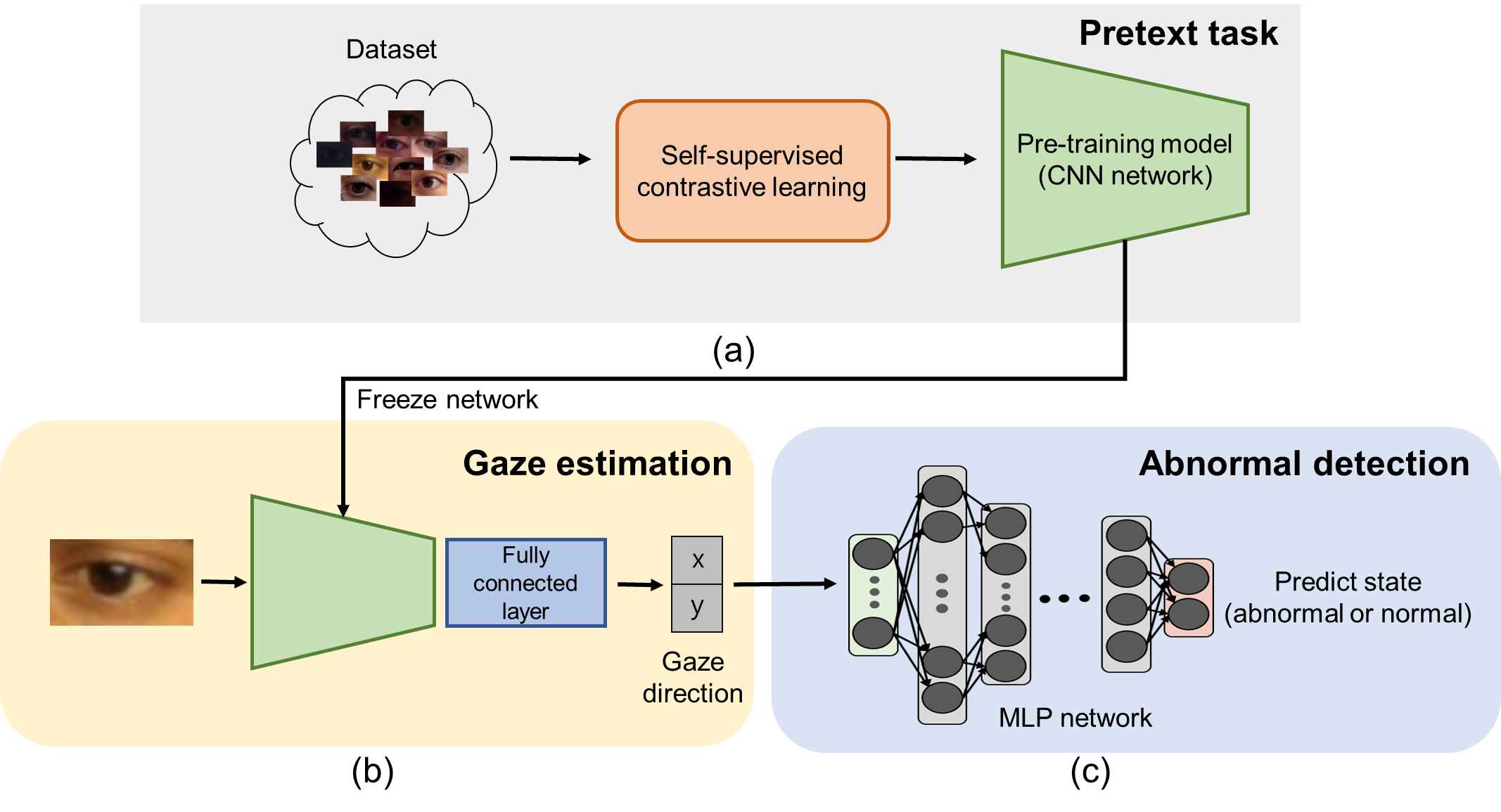}
\caption{The framework overview of our method. (a) The pretext task is performed through self-supervised contrastive learning. (b) Freeze the weights of the model from the pretext task and train the gaze estimation algorithm using this model. (c) The value inferred by the gaze estimation algorithm passes the MLP model and infer the state.} \label{fig1}
\end{figure}

The human gaze is considered one of the most important factors in behavioral analysis. Various analyses such as intent, emotion, and communication are possible through the eye's gaze. Since human eyes contain a lot of information, many applications use this information : mobile phone scenarios\cite{ref_5,ref_6}, virtual reality\cite{ref_7,ref_8}, content creation\cite{ref_9}, gaming\cite{ref_10}, health care\cite{ref_11,ref_12}, human robotic interaction (HRI)\cite{ref_13,ref_14,ref_50}, human action recognition (HAR)\cite{ref_49}, human computer interaction\cite{ref_15,ref_16,ref_52,ref_53}. 

There are two representative gaze estimation methods using deep learning. One is the model-based method and the other is the appearance-based method. The model-based method\cite{ref_17,ref_18,ref_19,ref_20} estimates the gaze value of the eyes with high accuracy, as it estimates the gaze using the 3D model of the eye. However, estimating gaze requires a variety of information, such as the eye's landmark, eye radius, and pupil size, and when one information is missing, the exact gaze estimation cannot be estimated. Due to these problems, the method is inappropriate for practical application The appearance-based methods\cite{ref_21,ref_22,ref_23,ref_24} pass the image of the eye through a convolutional neural network (CNN) to estimate the gaze value directly. This method does not require much information to estimate gaze, and can be estimated only with an eye image. Because the appearance method estimates gaze using a single image, it has a disadvantage of lower accuracy than the model-based method, but it is suitable for practical applications.

\begin{figure}
\centering
\includegraphics[width=0.9\textwidth]{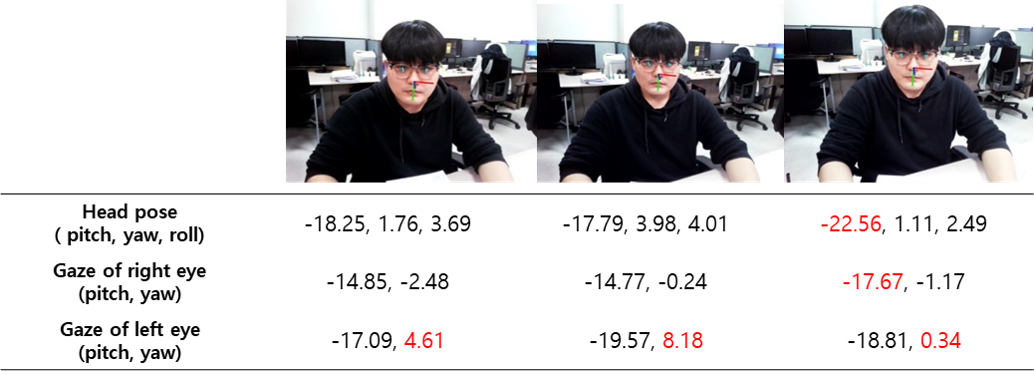}
\caption{Results of the appearance-based gaze estimation algorithm. It shows that there is a large difference in the inferred results even when looking at the same point.} \label{fig2}
\end{figure}

In this paper, we present a novel method that contributes to video conferencing solutions using gaze estimation algorithms. Fig. 1 shows the overall framework of our method. We use gaze estimation\cite{ref_22} of the appearance-based method because we aim to apply it to practical applications. We use the estimated values from the gaze estimation algorithm to check the normal, abnormal states of video conferencing users, where normal state means when video conferencing users stare at the monitor and abnormal means the opposite. 

The gaze estimation algorithm estimates a total of 7 values (yaw and pitch in the eyes, yaw and pitch in the head, distance from the camera). However, as mentioned earlier, it is very difficult to distinguish between normal and abnormal states by setting a threshold from these values because the appearance-based method estimates values of low accuracy. Fig. 2 shows the results of the gaze estimation algorithm. The user in the picture shows a large difference in estimates despite staring at the same point on the monitor. To address these problems, we use Multi-Layer Perceptron (MLP) to learn estimated values from gaze estimation algorithms to determine normal, abnormal states. We construct a new dataset by collecting seven estimated values from the gaze estimation algorithm to apply this method. To create sophisticated dataset, we first find the optimal network through various gaze estimation network experiments. Next, to solve the low accuracy problem of the appearance-based method, we apply the self-supervised method\cite{ref_32} to pre-train the model well to learn the representation of eye images and use this model to conduct transfer learning on the gaze estimation algorithm. We use this trained gaze estimation algorithm to create data for two states (normal, abnormal) and train these data on the MLP to determine the state of video conferencing users. Our contributions are as follows.

\begin{enumerate}
\item We find and apply optimal networks for gaze estimation through various network experiments and next apply a self-supervised method to complement the low accuracy problem of the appearance-based method.
\item For abnormal detection, we provide a new dataset for the seven values (eyes yaw, eyes pitch, head yaw, head pitch, distance from camera) of gaze estimation.
\item We propose a novel method by learning a new dataset using deep learning-based Multi Layer Perceptron (MLP) to determine normal and abnormal states.
\end{enumerate}

\section{Related Work}
Our research is related to self-supervised, gaze estimation. We would like to briefly address these points in this chapter. 

\subsection{Self-Supervised Learning}
Self-supervised learning is a subset of the unsupervised method. Recently, self-supervised representation learning methods based on deep learning have been developed for various domains. \cite{ref_25,ref_26} is a self-supervised method using a natural language domain, and \cite{ref_27,ref_28,ref_29} is a method using an image domain. Methods for using image domains are used by pixel prediction approaches to learn embeddings. However, a more effective method than this is to replace the dense per-pixel predictive loss with a loss of lower-dimensional representation space. Thus, self-supervised methods representing state-of-the-art performance use contrastive learning methods\cite{ref_30,ref_31,ref_32,ref_33} to apply this paradigm. 

Contrastive learning is familiar with losses based on metric distance learning or triplets\cite{ref_34,ref_35} and these losses are used to train robust representation. The difference between triplet losses and self-supervised constant losses is the number of positive and negative pairs used per data point. Triplet losses use exactly one positive pair from the same class and one negative pair from the other class for learning. Self-supervised contrastive losses similarly use a single positive pair, and positive pairs are selected through co-occurrence\cite{ref_30} or data augmentation\cite{ref_32}. The most noticeable difference is the use of a large number of negative pairs. Assuming that using many negative samples yields a low probability of false-negative, increasing the number of negative samples improves the performance of the representation. Typically, a self-supervised method pre-train a model using unlabeled data as pretext tasks, and then performs transfer learning on the downstream task

In order to complement the low accuracy problem of the appearance-based method of gaze estimation, we use the self-supervised contrastive learning method \cite{ref_32} to improve the representation of the eye image well. Then, improve gaze estimation by learning the gaze estimation algorithm using the pretext model obtained from this self-supervised method

\subsection{Gaze Estimation}

The model-based gaze estimation calculates the gaze value by mapping the eye image to a 3d space. In this method, the center, radius values of the eyeball are obtained and calculated using the camera coordinate system to place them in 3d space\cite{ref_19,ref_20,ref_51}. The center value of the eyeball is determined through landmarks in 2D space, and then the gaze value is obtained by the 3D Geometry eye model\cite{ref_20}. The method estimates a gaze value of high accuracy, but requires multiple pieces of information to obtain the value and cannot extract the gaze value if one requirement is not met. For this reason, it is difficult to use in practical applications.

The appearance-based gaze estimation method is a representative gaze estimation method using deep learning. The method proceeds with gaze estimation by training on CNN networks using datasets that it is eye image with gaze values labels. Since eye images are inserted into CNN networks to directly estimate gaze, they are less demanding and faster than model-based methods, making them suitable for practical applications\cite{ref_21,ref_23,ref_38,ref_39}. However, this method usually requires a lot of data and does not take into account the inter-subject variation in eye appearance. To address these challenges, studies have also produced new sophisticated datasets\cite{ref_5,ref_40,ref_41} and applied them to the appearance-based method, but they have not solved this problem clearly. 

We use an appearance-based method that is appropriate to applications, albeit with low accuracy, to determine the user's condition in video conferencing. We proceeded to improve the performance of the gaze estimation algorithm by applying self-supervised methods\cite{ref_32} to solve the low accuracy problem. We also applied Kalman Filter to minimize the jitter of the gaze values and using this created gaze estimation algorithm, we produced a sophisticated new dataset for video conferencing solutions.

\subsection{Dataset}
$\textbf{MpIIGaze}$ This dataset consists of 3,000 eye images each (1500 left and 1500 right) of 15 people. The eye images are provided in 60 x 36 pixels and are already frontalized depending on the values of yaw and pitch in the head pose. The gaze is labeled yaw, pitch using a coordinate system such as a head pose.

\section{Method}

This section introduces the overall methods. We first deal with self-supervised contrastive learning (Sec. 3.1). And then, section 3. 2 describe the gaze estimation method and describe the abnormal detection method (Sec. 3. 3)

\subsection{Self-Supervised Contrastive Learning}

\begin{figure}
\centering
\includegraphics[width=1\textwidth]{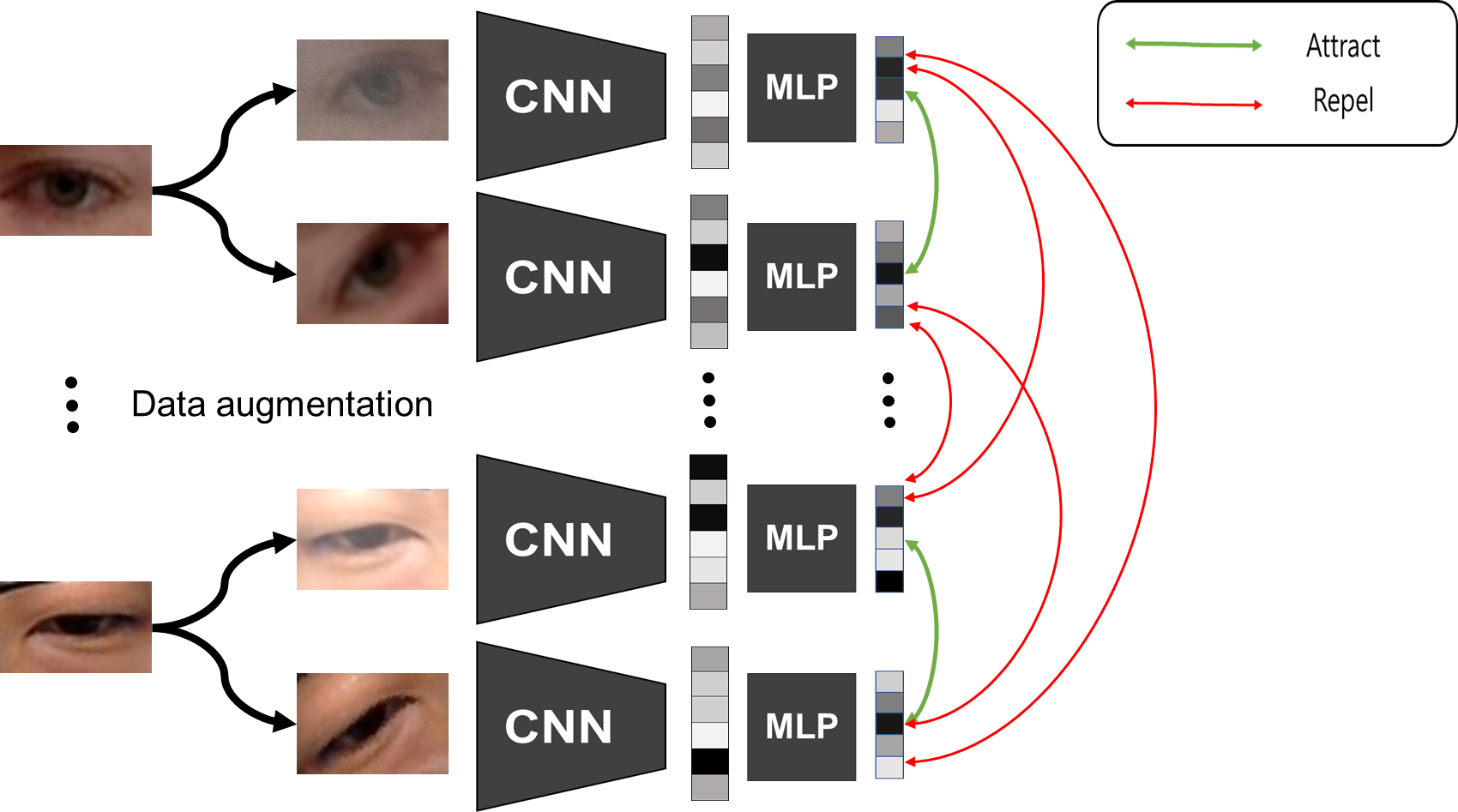}
\caption{Self-supervised contrastive learning using the gaze dataset\cite{ref_22}. The green line means the connection of positive pairs, and the red line means the connection of negative pairs.} \label{fig3}
\end{figure}

This section deals with self-supervised contrastive methods. Contrastive learning method, positive and negative pairs are selected for data, and positive pairs are pulled to become closer to each other and negative pairs move away from each other. Through this method, representations of the same class are trained to be close to each other, and representations of different classes are trained to be farther away from each other so that the boundary can be distinguished well. \cite{ref_32} is a representative self-supervised contrastive method and is called SimCLR. This method generates two images through augmentation, and the two images are defined as a positive pair. Then, from another image, two images are created in the same way, and these images are defined as negative pairs to proceed with contrastive learning. \cite{ref_32} increases the performance of contrastive learning by increasing the number of negative pairs using an augmentation method. 

We use this self-supervised method on the MPIIGaze dataset. The number of datasets consists of a total of 45,000 eye image data, and only four augmentation methods (color distort, rotation, gaussian noise, gaussian blur) are applied to maintain the overall shape of the eye image. Fig. 3 shows SimCLR using gaze dataset. Each eye image is randomly augmented to generate two image pairs and construct positive pair and negative pair. Then, self-supervised contrastive learning is performed using these data samples.

\subsubsection{Contrastive loss.}
When the total number of images is N, 2 (N-1) negative samples are configured for one data, and the pretext task learning using the contrastive learning method is performed using these samples. Eq. (1) is a loss function used in training, and (i, j) is defined as a positive pair.

\begin{equation}
    L_{i,j} = -\log\frac{exp(sim(z_{i}, z_{j})/\tau)}{\sum_{k=1}^{2N}\mathbf{1_{[k\ne i]}}exp(sim(z_{i}, z_{k})/\tau)}
\end{equation}\\

where $\mathbf{1_{[k\ne i]}}$ means one indicator function, and $\tau$ means a temperature parameter. ($z_{i}$, $z_{j}$) represents a positive pair and ($z_{i}$, $z_{k}$) represents a negative pair. Self-supervised contrastive learning is performed through this loss function.

\subsection{Gaze Estimation}

Fig. 4 shows the gaze estimation architecture. We used the gaze estimation method of \cite{ref_22} as a baseline. To perform gaze estimation algorithm training, the model trained by self-supervised contrastive learning is used as the backbone and fully connected layers are added later to perform fine-tuning. The eye image passes the CNN and performs a final gaze regression by adding the head pose value (yaw and pitch) in the fully connected layer.

\begin{figure}
\centering
\includegraphics[width=0.8\textwidth]{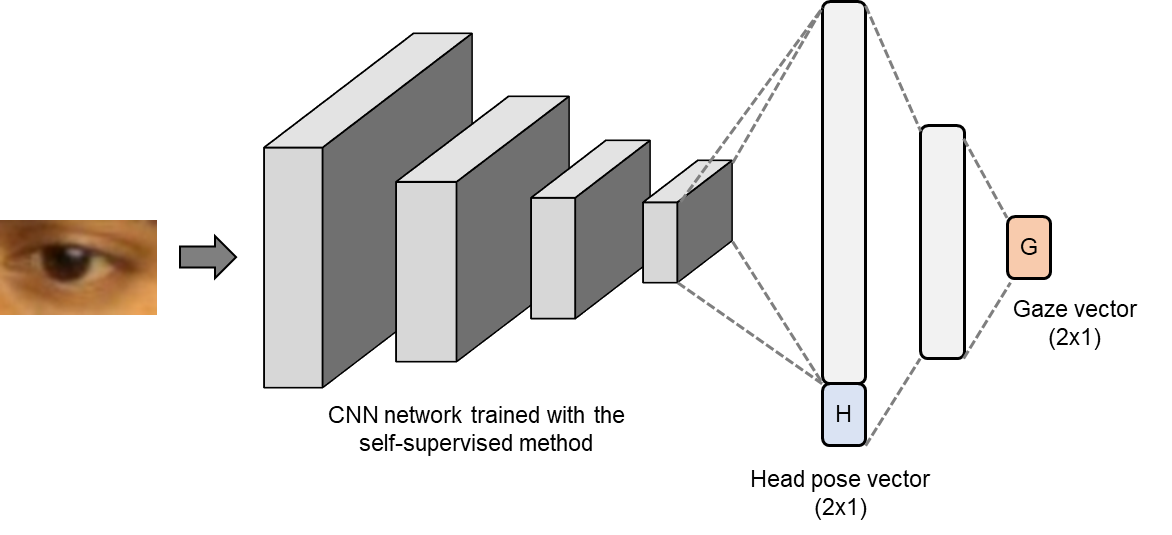}
\caption{The gaze estimation architecture. The appearance-based gaze estimation algorithm is learned using the model learned from the self-supervised contrastive method as the backbone.} \label{fig4}
\end{figure}

\subsubsection{Gaze loss.}
Gaze estimation model is learned using L2 loss (see Eq. 2).
\begin{equation}
    L(I) = \frac{1}{N}\sum_{n=1}^N\parallel g_g - g_p \parallel_2.
\end{equation}

\noindent $g_p$ is the predicated gaze from image I and $g_g$ is ground truth.

\subsubsection{Regularization of gaze value.}

Since appearance-based estimation directly estimates gaze values from images, jitter problems arise in the estimation values. Because of this problem, high-quality data cannot be obtained when creating a data set for abnormal detection. To solve this problem, we alleviated the jitter problem by applying the kalman smoother \cite{ref_42} to the yaw and pitch values of the eyes estimated from the gaze estimation algorithm.

\subsection{Abnormal Detection}

For abnormal detection, we have created a new dataset. Fig. 5 shows the setting environment for creating a new dataset and data appearance. We constructed a data set using the 7 values (yaw and pitch in the eyes, yaw and pitch in the head, distance from the camera) estimated by the gaze estimation algorithm, and labeled as 1 when looking at the normal area of the monitor and was labeled as 0 when gazing into the abnormal area. We produced a total of 700K data through these environment settings. The generated data is trained through Multi Layer Perceptron (MLP) and infers two states by utilizing the trained model.

\begin{figure}
\centering
\includegraphics[width=0.8\textwidth]{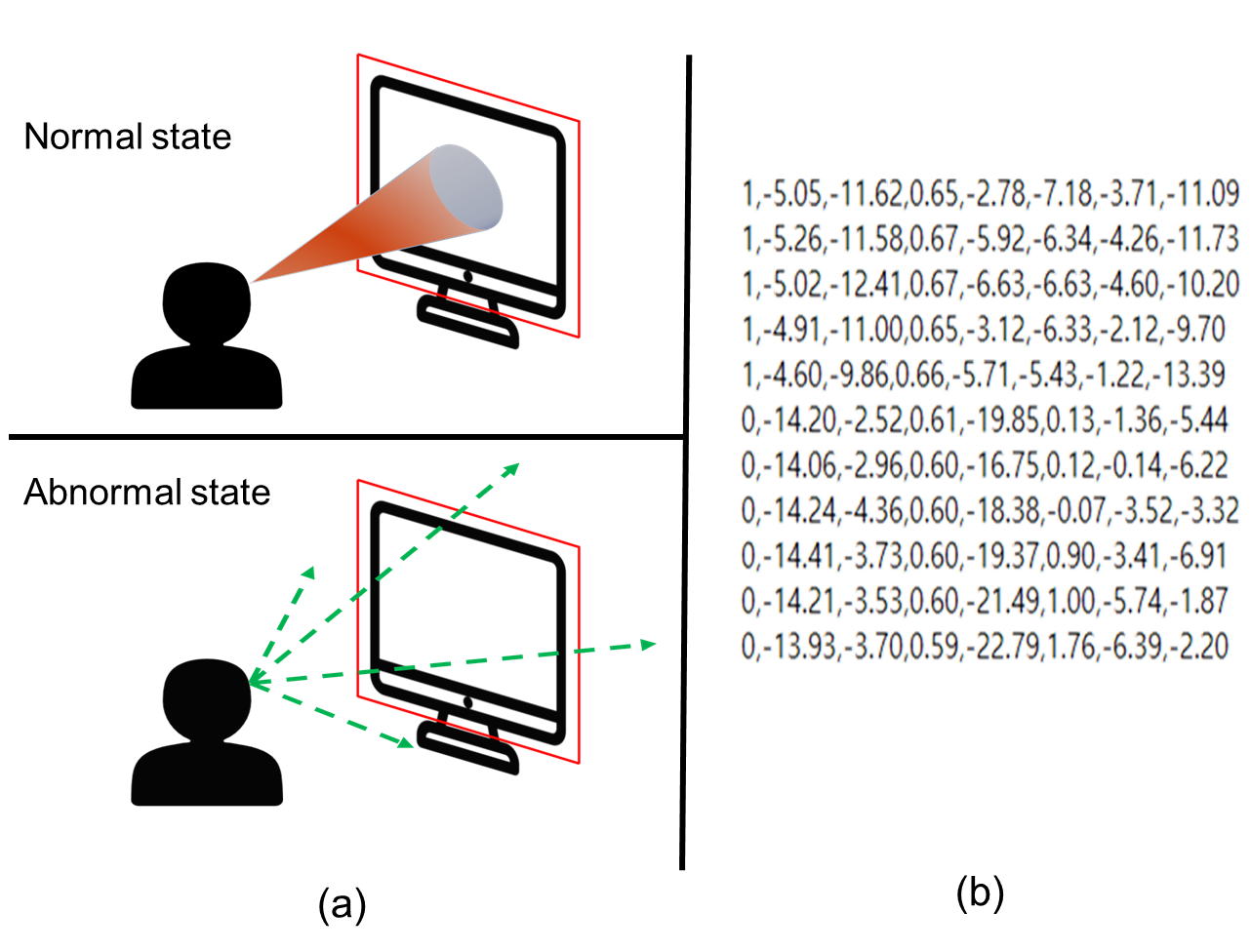}
\caption{The setting environment for data creation and the appearance of data created. (a) It shows the normal state and the abnormal state, which of if the gaze enters the red boundary, it is a normal state, otherwise, it is in an abnormal state. (b) It shows the shape of the data created using the estimated value according to the state from the gaze estimation algorithm, which 0 means abnormal and means normal.
} \label{fig5}
\end{figure}

\subsubsection{Cross entropy loss.}
We train the MLP model using cross-entropy loss (see Eq. 3).
\begin{equation}
   L_{CE} = -\frac{1}{N}\sum_{i=1}^n\sum_{c=1}^C L_{ic}\log(P_{ic})
\end{equation}

where n is the number of data, C is the number of classes, $L_{ic}$ is the ground truth label and And $P_{ic}$ is Softmax probability for $I^{th}$ class

\section{Experiment and Result}

In this section, we present the experimental results of our proposed methods. We pre-trained the self-supervised method using the MPIIGaze dataset\cite{ref_22}, performed the pretext task, and fine-tuned the gaze estimation algorithm using the trained network. Next, a new dataset was constructed by collecting the estimated values from the learned gaze estimation algorithm, and this dataset was trained using the MLP model to perform abnormal detection. Section 4.1 demonstrates the robustness of our method by comparing before and after applying self-supervised to gaze estimation. Section 4.2 shows how to perform abnormal detection by training the MLP model using the dataset we created.

\subsection{Gaze Estimation with Self-Supervised Method}

\begin{table}[h]
\begin{center}
\begin{tabular}{|c|c|c|c|}
\hline
\multicolumn{2}{|c|}{Baseline method} & \multicolumn{2}{|c|}{Our method} \\
\hline\hline
Network & Loss & Network & Loss\\\hline
LeNet & 3.53 & LeNet & 3.23($\downarrow$ 0.30)\\
ResNet18 & 3.82 & ResNet18 & 3.51($\downarrow$ 0.31)\\
SqueezeNet & 3.60 & SqueezeNet & 3.26($\downarrow$ 0.34)\\
MobileNetv2 & 4.5 & MobileNetv2 & 3.62($\downarrow$ 0.88)\\
ShuffleNet & 4.23 & ShuffleNet & 3.52($\downarrow$ 0.71)\\
\hline
\end{tabular}
\end{center}
\caption{Results compared with the baseline. We demonstrate the robustness of our method by confirming that the performance of all networks used in the experiment is improved compared to when the baseline method is used.
}
\end{table}

\subsubsection{Self supervised training detail.}

We conducted pre-training on the self-supervised method[32] with a total of 45,000 eye images. We used a total of four augmentations (color distort, rotation, gaussian noise, gaussian blur) to maintain the shape of the eye image. We used two Nvidia 1080ti and an inter core i7 CPU to train at 10000 epochs with 1024 batch size.

\subsubsection{Gaze estimation training detail.}

We train the gaze estimation algorithm by adding some fully connected layers after freezing the model weights obtained from self-supervised learning. The batch size was set to 32, and the learning rate was decreased by $10^{-1}$ every 200 epochs starting at $10^{-4}$. To prevent overfitting, a weight decay of $10^{-4}$ was applied and an SGD (Stochastic Gradient Descent) optimizer\cite{ref_43} was used. 

\subsubsection{Evaluation result.}

We trained the gaze estimation algorithm by applying the self-supervised method using a total of 5 CNN networks\cite{ref_23,ref_45,ref_46,ref_47,ref_48} and compared the performance when not applied. Tab. 1 shows the comparison results.

From the results in the table, our method decreased the loss by 0.5 on average compared to the baseline method. We use these experimental results to prove the robustness of our new gaze estimation method.

\subsection{Abnormal Detection}

\begin{figure}
\centering
\includegraphics[width=.9\textwidth]{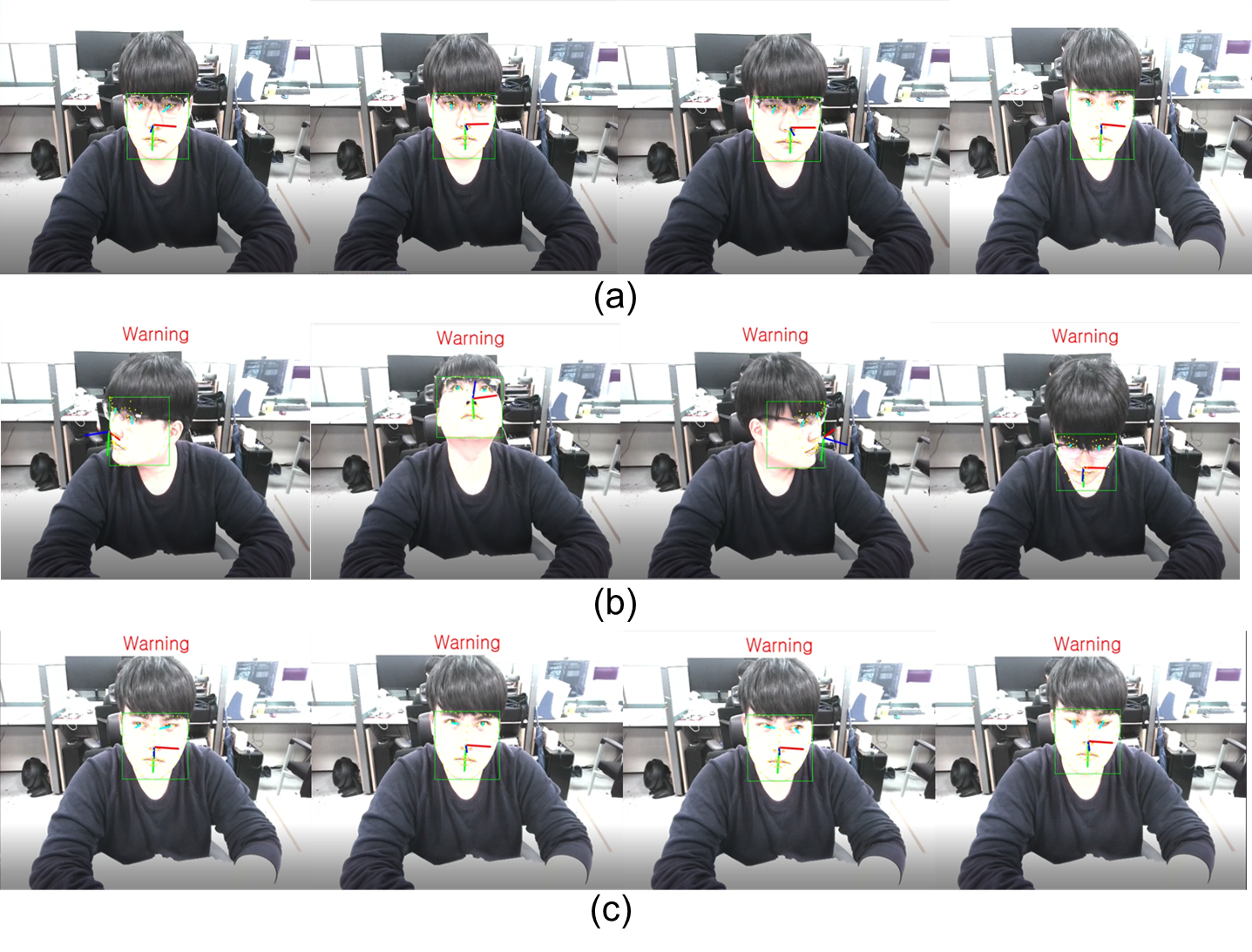}
\caption{Results of our abnormal detection method. (a) The normal state looking at the monitor. (b) The abnormal state in which the head is turned and the gaze is not directed to the monitor. (c) The abnormal state with the head facing the monitor but the gaze looking out of the monitor} \label{fig6}
\end{figure}

\subsubsection{Dataset configuration.}
We created a dataset by learning the gaze estimation algorithm using LeNet\cite{ref_23}, which has the best performance in the previous experiment. We created a total of 700K dataset and consisted of 500K training data, 200K validation data, and 100K test data.

\subsubsection{MLP training detail for abnormal detection.}

We set the batch size to 128 and the learning rate to start at $10^{-2}$ and decrease by $10^{-1}$ every 400 epochs. We trained using the Adam optimizer\cite{ref_44}, constructed an MLP with a total of three and four fully connected layers, and trained the dataset created using this model.

\subsubsection{Evaluation result.}

We achieved an accuracy of 91\% on the test data set with our constructed MPL model. The trained MLP model was attached to the gaze estimation algorithm to determine the status of video conferencing users. Fig. 6 shows the results of our proposed abnormal detection method.

Using our proposed model, we can display a warning message when a user turns their head or looks away from the monitor. Through these results, our method shows the potential for application to video conferencing solutions.

\section{Conclusion}

We design and experiment with a novel abnormal detection method that can be applied to video conferencing solutions using gaze estimation algorithms. At first, we tried abnormal detection by giving a threshold using gaze and head pose. However, the gaze estimation's inference values were not suitable for this method due to the large variation of the values even in the slightest movement. Even if I stare at the same point, the camera and the person's position change greatly depending on the movement of the position. So we applied deep learning methods to solve these problems. We extracted seven values from the gaze estimation method, labeled them abnormal and normal, and trained them on the MLP model. This method detected abnormal behavior better than the threshold method. However, this method still has many things to be solved. First of all, this experiment produced data from one subject. However, considering the various shapes of a person's eye shape, it is necessary to produce data from various subjects. Since there are also differences in the predicted values depending on the size of the monitor and the position of the camera, there are still many improvements to be made in order to be applied to practical applications. If this is solved, our method is thought to be of great help to the video conferencing solution.

%
%
\newpage

%

\end{document}